\documentclass[10pt,journal,final]{IEEEtran}

\usepackage[utf8]{inputenc} 
\usepackage[T1]{fontenc}    
\usepackage{url}            
\usepackage{booktabs}       
\usepackage{amsfonts}       
\usepackage{nicefrac}       
\usepackage{microtype}      
\usepackage{amsmath,amssymb,marvosym,graphicx,bm}
\usepackage[colorlinks=true,citecolor=blue]{hyperref}
\usepackage{subfigure}

\def\eg{{\em e.g.}}
\def\etal{\emph{et al.}}
\def\ie{{\em i.e.}}

\makeatletter
\renewcommand\normalsize{%
  \@setfontsize\normalsize\@xpt\@xiipt
   \abovedisplayskip 8\p@ \@plus2\p@ \@minus5\p@
   \abovedisplayshortskip \z@ \@plus3\p@
   \belowdisplayshortskip 6\p@ \@plus3\p@ \@minus3\p@
   \belowdisplayskip \abovedisplayskip
   \let\@listi\@listI}
\makeatother

\begin{document}

\title{Align Deep Features for Oriented Object Detection}

\author{Jiaming~Han,
        \and Jian~Ding, 
        \and Jie Li,
        \and Gui-Song~Xia,~\IEEEmembership{Senior~Member,~IEEE}
\thanks{The study in this article is funded by the National Natural Science Foundation of China (NSFC) under the grant contracts No.61922065, No.61771350, No.41820104006, and No.61871299. It is also partially funded by the Science and Technology Major Project of Hubei Province (Next-Generation AI Technologies) under Grant 2019AEA170, and the Shanghai Aerospace Science and Technology Innovation Project (SAST2019-094).}
\thanks{J. Han, J. Ding are with the State Key Lab. LIESMARS and the Institute of Artificial Intelligence, Wuhan University, Wuhan, 430079, China.}
\thanks{J. Li is with Shanghai Aerospace Electronic Technology Institute, Shanghai, China. E-mail: {\em trackerdsp@163.com}.}
\thanks{G.-S. Xia is with the National Engineering Research Center for Multimedia Software, School of Computer Science and Institute of Artificial Intelligence, and also the State Key Lab. LIESMARS, Wuhan University, Wuhan, 430072 China. E-mail: {\em guisong.xia@whu.edu.cn}.}
\thanks{J. Han and J. Ding are equally contributed to this work.}
\thanks{Corresponding author: Gui-Song Xia ({\em guisong.xia@whu.edu.cn}).}
}
\maketitle

\begin{abstract}
The past decade has witnessed significant progress on detecting objects in aerial images that are often distributed with large scale variations and arbitrary orientations. However most of existing methods rely on heuristically defined anchors with different scales, angles and aspect ratios and usually suffer from severe misalignment between anchor boxes and axis-aligned convolutional features, which leads to the common inconsistency between the classification score and localization accuracy. To address this issue, we propose a {\em Single-shot Alignment Network} (S$^2$A-Net) consisting of two modules: a Feature Alignment Module (FAM) and an Oriented Detection Module (ODM). The FAM can generate high-quality anchors with an Anchor Refinement Network and adaptively align the convolutional features according to the anchor boxes with a novel Alignment Convolution. The ODM first adopts active rotating filters to encode the orientation information and then produces orientation-sensitive and orientation-invariant features to alleviate the inconsistency between classification score and localization accuracy. Besides, we further explore the approach to detect objects in large-size images, which leads to a better trade-off between speed and accuracy. Extensive experiments demonstrate that our method can achieve state-of-the-art performance on two commonly used aerial objects datasets (\ie, DOTA and HRSC2016) while keeping high efficiency\footnote{The code is available at~\url{https://github.com/csuhan/s2anet}}.
\end{abstract}


\section{Introduction}
\label{sec:intro}
\IEEEPARstart{O}{bject} detection in aerial images aims at identifying the locations and categories of objects of interest (\eg, planes, ships, vehicles). With the framework of deep convolutional neural networks, object detection in aerial images (ODAI) has made significant progress in recent years~\cite{cheng2016ricnn,liu2016ship,xia2018dota,ding2018transformer,zhang2019cad,wang2020centermap,xu2019gliding}, where most of existing methods are devoted to cope with the challenges raised by the large scale variations and arbitrary orientations of crowded objects in aerial images.

To achieve better detection performance, most state-of-the-art aerial object detectors~\cite{ding2018transformer, zhang2019cad, yang2019scrdet, xu2019gliding} rely on the complex R-CNN~\cite{girshick2014rich} frameworks, which consist of two parts: a Region Proposal Network (RPN) and an R-CNN detection head. 
In a general pipeline, RPN is used to generate high-quality Region of Interests (RoIs) from horizontal anchors, then an RoI Pooling operator is adopted to extract accurate features from RoIs. Finally, R-CNN is employed to regress the bounding boxes and classify them into different categories. 
\footnotetext[1]{\url{https://github.com/dingjiansw101/AerialDetection}.}
However, it is worth noticing that horizontal RoIs often result in severe misalignment between bounding boxes and oriented objects~\cite{ding2018transformer,xia2018dota}. For example, a horizontal RoI usually contains several instances due to oriented and densely packed objects in aerial images. 
A natural solution is employing oriented bounding boxes as anchors to alleviate this issue~\cite{liu2016ship, xia2018dota}. As a consequence, well-designed anchors with different angles, scales and aspect ratios are required, which however leads to massive computations and memory footprint. 
Recently, RoI Transformer~\cite{ding2018transformer} was proposed to address this issue by transforming horizontal RoIs into rotated RoIs, avoiding a large number of anchors, but it still needs heuristically defined anchors and complex RoI operation.

\begin{figure}[htp!]
  \centering
  \subfigure[The misalignment issue and our solution]{
  \includegraphics[width= 0.87\linewidth]{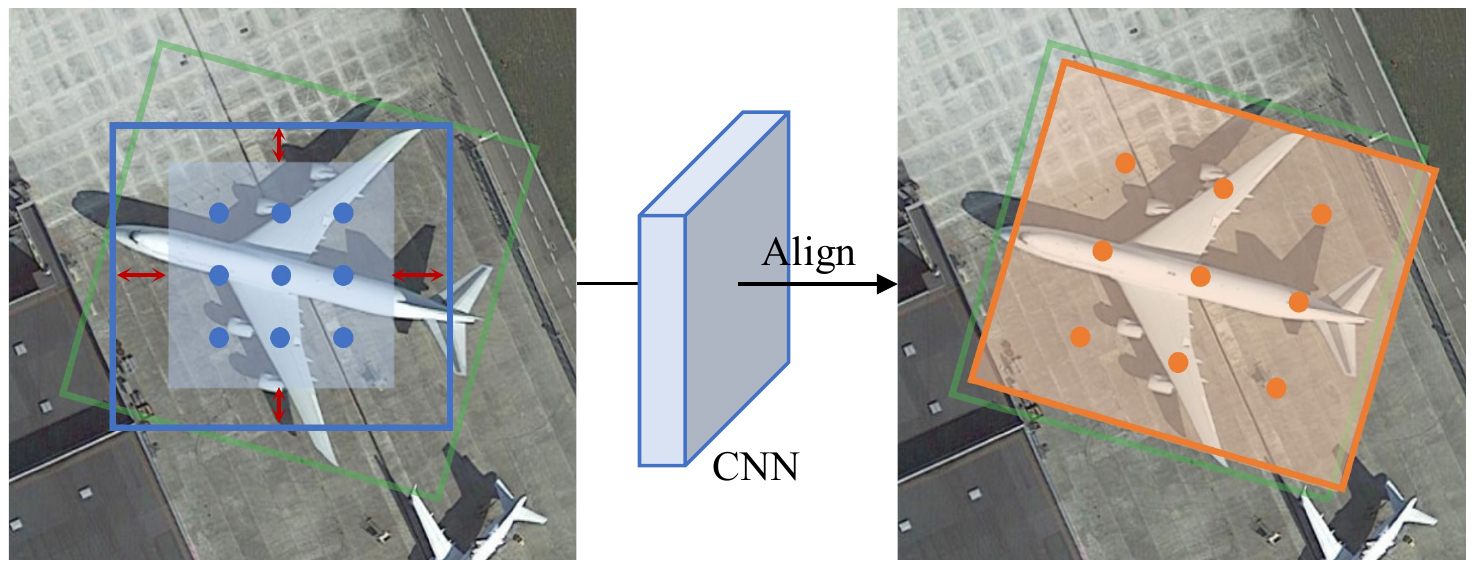}
  }
  \subfigure[Speed vs. accuracy (mAP) on DOTA]{
  \includegraphics[width= 0.83\linewidth]{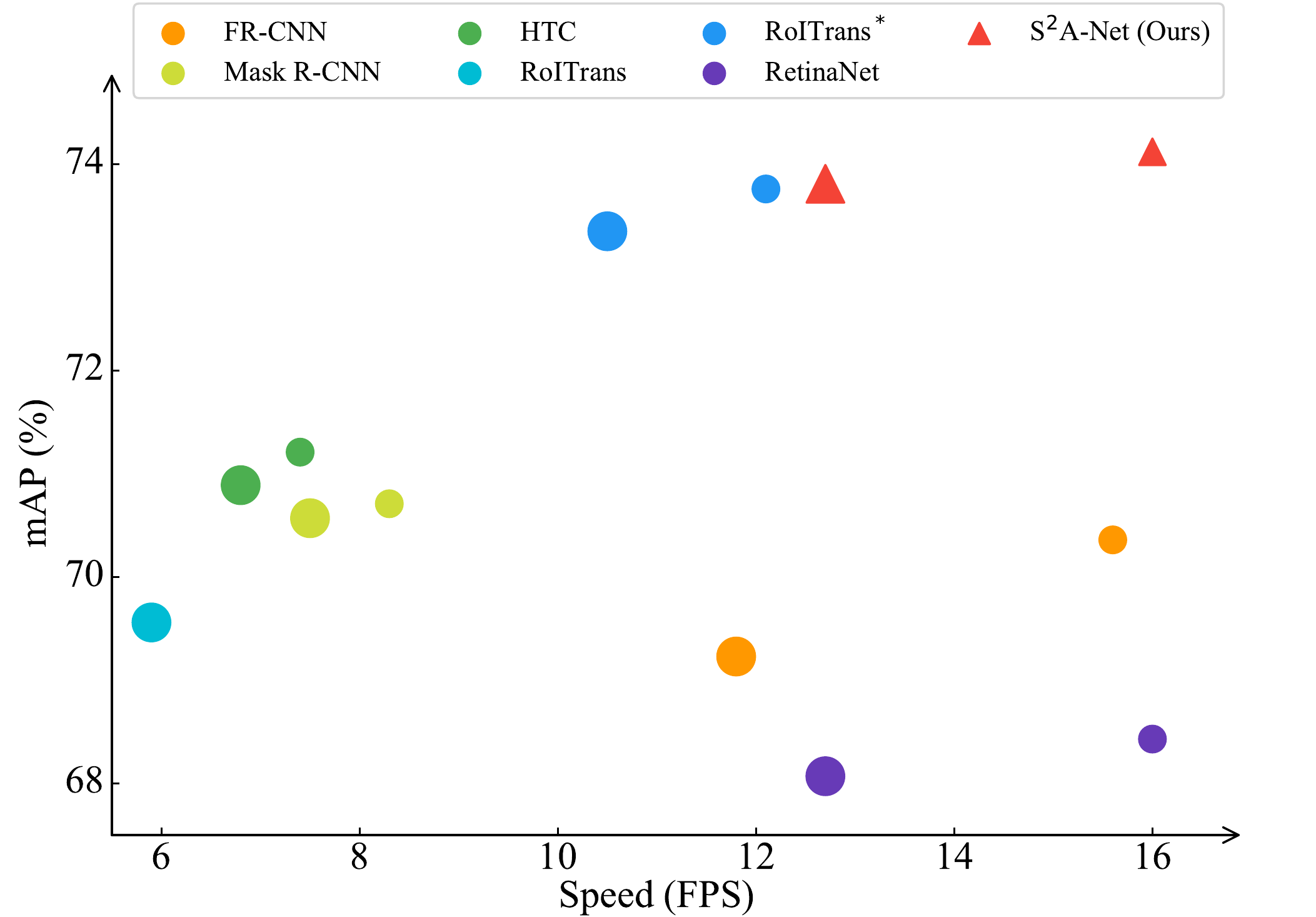}
  }
    \vspace{-2mm}
    \caption{{\bf (a)} {The misalignment (red arrows) between an anchor box (blue bounding box) and convolutional features (light blue rectangle)}. To alleviate this issue, we first refine the initial anchor into a rotated one (orange bounding box), and then adjust the feature sampling locations (orange points) with the guide of the refined anchor box to extract aligned deep features. The green box denotes the ground truth. {\bf (b)} Performance comparisons of different methods under the same settings: ResNet50 (in small markers) and ResNet101 (in 
   big markers) backbones, $1024\times1024$ input size of images, without data augmentation. Faster R-CNN (FR-CNN)~\cite{ren2017faster}, Mask R-CNN~\cite{he2017maskrcnn}, RetinaNet~\cite{lin2017focal}, Hybird Task Cascade (HTC)~\cite{chen2019htc} and RoI Transformer (RoITrans)~\cite{ding2018transformer} are tested. The speed of all methods is reported on the V100 GPU in terms of Frames Per Second (FPS). Note that Mask R-CNN, HTC and RoITrans are tested based on the \texttt{AerialDetection}\protect\footnotemark[1] project. RoITrans$^*$ indicates an official re-implementation.}
    \label{fig:alignment_fig}
   \vspace{-4mm}
  \end{figure}
  
In contrast with R-CNN based detectors, one-stage detectors regress the bounding boxes and classify them directly with regular and densely sampling anchors. This architecture enjoys high computational efficiency but often lags behind in accuracy~\cite{xia2018dota}. As shown in Fig.~\ref{fig:alignment_fig} (a), we argue that severe misalignment in one-stage detectors matters: 
\begin{itemize}
\item  [-]Heuristically defined anchors are with low-quality and cannot cover the objects, leading a misalignment between objects and anchors. For example, the aspect ratio of a bridge usually ranges from $1/3$ to $1/30$, and only a few or even no anchors can be assigned to it. This misalignment usually aggravates the foreground-background class imbalance and hinders the performance.
\item [-]The convolutional features from the backbone network are usually axis-aligned with fixed receptive field, while objects in aerial images are distributed with arbitrary orientations and variant appearances. Even an anchor box is assigned to an instance with high confidence (\ie, Intersection over Union (IoU)), there is still a misalignment between anchor boxes and convolutional features. In other words, the corresponding feature of an anchor box is hard to represent the whole object to some extent. As a result, the final classification score can not accurately reflect the localization accuracy, which also hinders the detection performance in post-processing phases (\eg, non-maximum suppression (NMS)).
\end{itemize} 

To address these issues in one-stage detectors, we propose a {\em Single-Shot Alignment Network} (S$^2$A-Net) which consists of two modules: a Feature Alignment Module (FAM) and an Oriented Detection Module (ODM). The FAM can generate high-quality anchors with an Anchor Refinement Network (ARN) and adaptively align the feature according to the corresponding anchor boxes (Fig~\ref{fig:alignment_fig}(a)) with an Alignment Convolution (AlignConv). Different from other methods with densely sampling anchors, we employ only one squared anchor for each location in the feature map, and the ARN refines them into high-quality rotated anchors. Then the AlignConv, a variant of convolution, adaptively aligns the feature according to the shapes, sizes and orientations of its corresponding anchors. In the ODM, we first adopt active rotating filters (ARF)~\cite{zhou2017orn} to encode the orientation information and produce orientation-sensitive features, and then extract orientation-invariant features by pooling the orientation-sensitive features. Finally, we feed the features into a regression sub-network and a classification sub-network to yield the final predictions. Besides, we also explore the approach to detect objects on large-size images (\eg, $4000\times4000$) rather than on chip images, which significantly reduces the overall inference time with negligible loss of accuracy. Extensive experiments on commonly used datasets, \ie, DOTA~\cite{xia2018dota} and HRSC2016~\cite{liu2017hrsc2016}, demonstrate that our proposed method can achieve state-of-the-art performance while keeping high efficiency, see Fig~\ref{fig:alignment_fig}~(b). 

Our main contributions are summarized as follows:
\begin{itemize}
  \item We propose a novel Alignment Convolution to alleviate the misalignment between axis-aligned convolutional features and arbitrary oriented objects in a fully convolutional way. Note AlignConv has negligible extra consuming time compared with standard convolution and can be embedded into many detectors with little modification.
  \item With the Alignment Convolution embedded, we design a light Single-Shot Alignment Network which enables us to generate high-quality anchors and aligned features for accurate object detection in aerial images.
  \item We report $79.42\%$ mAP on the oriented object detection task on the DOTA dataset, achieving the state-of-the-art in both speed and accuracy.
\end{itemize}

The rest of the paper is organized as follows. Section~\ref{sec:related_works} introduces the related works. 
Section~\ref{sec:methods} introduces the details of our proposed S$^2$A-Net. In Section~\ref{sec:experiments}, the experimental results and analysis are reported on challenging DOTA and HRSC2016 datasets. Finally, the conclusion is made in Section~\ref{sec:conclusion}.

\begin{figure*}[!t]
  \centering
  \includegraphics[width=.87\linewidth]{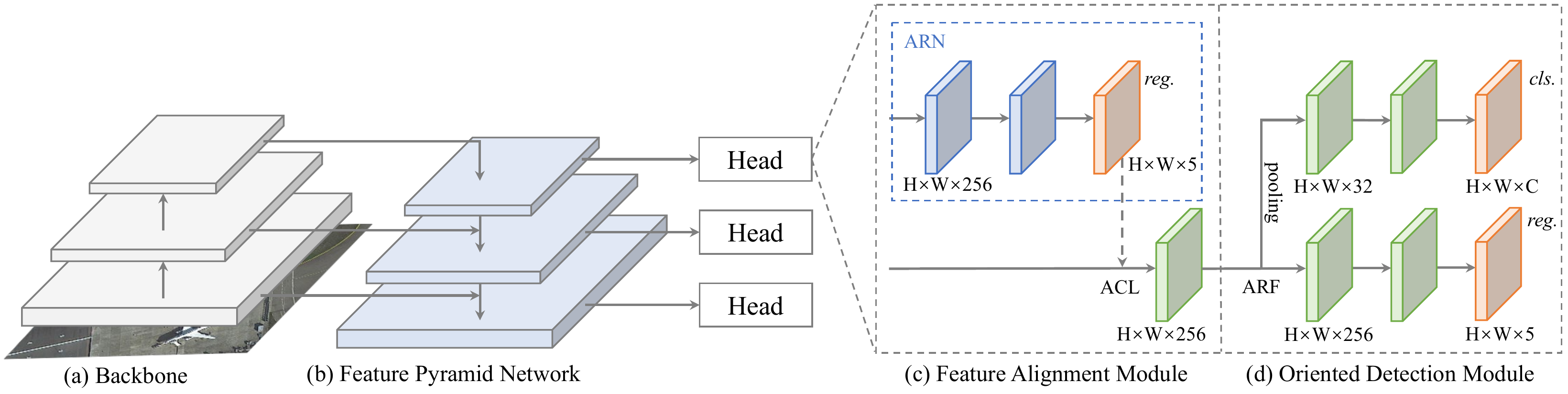}
  \vspace{-2mm}
  \caption{Architecture of the proposed S$^2$A-Net. S$^2$A-Net consists of a backbone network, a Feature Pyramid Network~\cite{lin2017focal}, a Feature Alignment Module (FAM) and an Oriented Detection Module (ODM). The FAM and ODM make up the detection head which is applied to each scale of the feature pyramid. In FAM, the Anchor Refinement Network (ARN) is proposed to generate high-quality rotated anchors. Then we feed the anchors and input features into the Alignment Convolution Layer (ACL) to extract aligned features. Note we only visualize the regression (\emph{reg.}) branch of ARN and ignore the classification (\emph{cls.}) branch for simplification.  In ODM, we first adopt active rotating filters (ARF)~\cite{zhou2017orn} to generate orientation-sensitive features, and pool the features to extract orientation-invariant features. Then the~\emph{cls.} branch and~\emph{reg.} branch are applied to produce the final detections.}
  \label{fig:S$^2$A-Net}
  \vspace{-3mm}
\end{figure*}

\section{Related Works}
\label{sec:related_works}
With the advance of machine learning, especially deep learning, object detection has made significant progress in recent years, which can be roughly divided into two groups: two-stage detectors and one-stage detectors. Two-stage detectors~\cite{girshick2014rich, girshick2015fast, ren2017faster, he2017maskrcnn} first generate a sparse set of RoIs in the first stage, and perform an RoI-wise bounding box regression and object classification in the second one. One-stage detectors, \eg, YOLO~\cite{redmon2016you} and SSD~\cite{liu2016ssd}, detect objects directly and do not require the RoI generation stage. Generally, the performance of one-stage detectors usually lag behind two-stage detectors due to extreme foreground-background class imbalance. To address this problem, the {\em Focal Loss}~\cite{lin2017focal} can be used, and anchor-free detectors~\cite{Law2018cornernet, zhou2019objects,yang2019reppoints} alternatively formulate object detection as a points detection problem to avoid complex computations related to anchors and usually run faster.

\vspace*{-3mm}
\subsection{Object Detection in Aerial Images} 
Objects in aerial images are often crowded, distribute with large scale variations and appear at arbitrary orientations. Generic object detection methods with horizontal anchors~\cite{xia2018dota} usually suffer from severe misalignment in such scenarios: one anchor/RoI may contain several instances. Some methods~\cite{liu2016ship,liu2017rotated,ma2018arbitrary} adopt rotated anchors with different angles, scales and aspect ratios to alleviate this issue, while involving heavy computations related to anchors (\eg, bounding box transform and ground truth matching). Ding~\etal~\cite{ding2018transformer} propose RoI Transformer to transform horizontal RoIs into rotated RoIs, which avoids a large number of anchors and alleviates the misalignment issue. However, it still needs heuristically defined anchors and complex RoI operations. Instead of employing rotated anchors, Xu~\etal~\cite{xu2019gliding} glide the vertex of the horizontal bounding box to accurately describe an oriented object. But the corresponding feature of a RoI is still horizontal and suffers from the misalignment issue. Recently proposed R$^3$Det~\cite{yang2019r3det} samples features from five locations (\eg, center and corners) of the corresponding anchor box and sum them up to re-encode the position information. In contrast with the above methods, the proposed S$^2$A-Net in this paper gets ride of heuristically defined anchors and can generate high-quality anchors by refining horizontal anchors into rotated anchors. Besides, the proposed FAM module enables to achieve feature alignment in a fully convolutional way.  

\vspace*{-3mm}
\subsection{Feature Alignment in Object Detection} 
Feature alignment usually refers to the alignment between convolution features and anchor boxes/RoIs, which is important for both two-stage and one-stage detectors. Detectors relying on misaligned features are hard to obtain accurate detections. In two-stage detectors, an RoI operator (\eg, RoIPooling~\cite{girshick2015fast}, RoIAlign~\cite{he2017maskrcnn} and Deformable RoIPooling~\cite{dai2017dcn}) is adopted to extract fixed-length features inside the RoIs which can approximately represent the location of objects. RoIPooling first divides an RoI into a grid of sub-regions and then max-pools each sub-region into the corresponding output grid cell. However, RoIPooling quantizes the floating-number boundary of an RoI into integer, which introduces misalignment between the RoI and the feature. To avoid the quantization of RoIPooling, RoIAlign adopts bilinear interpolation to compute the extract values at each sampling location in sub-regions, significantly boosting the performance of localization. Meanwhile, Deformable RoIPooling adds an offset to each sub-region of an RoI, enabling adaptive feature selection. However, the RoI operator usually involves massive region-wise operation, \eg, feature warping and feature interpolation, which becomes a bottleneck toward fast object detection. 

Recently, Guided Anchoring~\cite{wang2019region} tries to align features with the guide of anchor shapes. It learns an offset field from the anchor prediction map and then guides the Deformable Convolution (DeformConv) to extract aligned features. AlignDet~\cite{chen2019aligndet} designs an RoI Convolution to obtain the same effect as RoIAlign in one-stage detector. Both~\cite{wang2019region} and~\cite{chen2019aligndet} achieve feature alignment in a fully convolutional way and enjoy high efficiency. These methods work well for objects in nature images but often lose their performance when detecting objects that are oriented and densely packed in aerial images, although some of them (\eg, Rotated RoIPooling~\cite{ma2018arbitrary} and Rotated Position Sensitive RoIAlign~\cite{ding2018transformer}) have been adopted to achieve feature alignment in oriented object detection. Different from the aforementioned methods, our proposed method aims at alleviating the misalignment between axis-aligned convolutional features and arbitrary oriented objects, which adjusts the feature sampling locations with the guide of anchor boxes.

\vspace*{-3mm}
\subsection{Inconsistency between Regression and Classification}
An object detector usually consists of two parallel tasks: bounding-box regression and object classification, which share the same features from the backbone network. And the classification score is used to reflect the localization accuracy in a post-processing phase (\eg, NMS). However, as discussed in~\cite{Jiang2018iounet} and~\cite{wu2019rethinking}, there is a common inconsistency between classification score and localization accuracy. Detections with high classification scores may produce bounding boxes with low localization accuracy. While other nearby detections with high localization accuracy may be suppressed in the NMS step. To address this issue, IoU-Net~\cite{Jiang2018iounet} proposed to learn to predict the IoU of a detection as the localization confidence and then combine the classification score and localization confidence as the final probability of a detection. Double-Head R-CNN~\cite{wu2019rethinking} adopts different head architectures for different tasks, \ie, fully connected head for classification and convolution head for regression. In our methods, we aim to improve the classification score by extracting aligned features for each instance. Especially when detecting densely packed objects in aerial images, accurate features are important to robust classification and precise localization. Besides, as discussed in~\cite{wu2019rethinking}, shared features from the backbone are not suitable for both classification and localization. Inspired by~\cite{zhou2017orn} and~\cite{liao2018rotation}, we first adopt active rotating filters to encode the orientation information and then extract orientation-sensitive features and orientation-invariant features for regression and classification, respectively.

\begin{figure}[!t]
  \centering
  \includegraphics[width=0.8\linewidth]{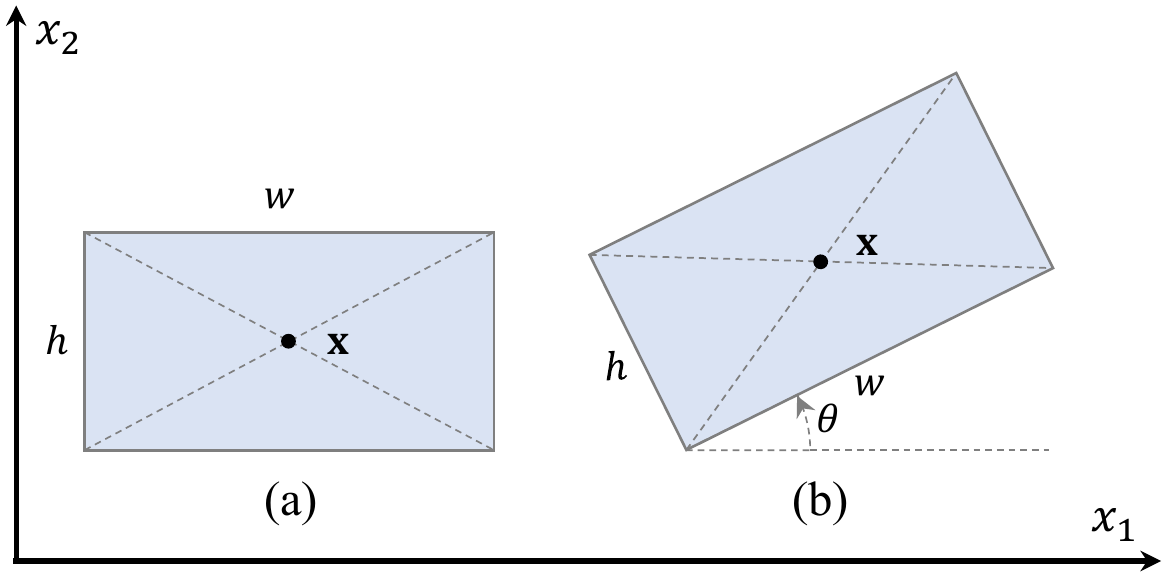}
   \vspace{-2mm}
  \caption{Two types of bounding box. {\bf (a)} Horizontal bounding box $\big\{\big( \mathbf x, w, h \big)\big\}$ with center point $\mathbf  x=(x_1,x_2)$, width $w$ and height $h$. {\bf (b)} Oriented bounding box $\big\{\big( \mathbf x, w, h, \theta \big)\big\}$. $\mathbf x$ denotes the center point. $w$ and $h$ represent the long side and short side of a bounding box, respectively. $\theta$ means the angle from the position direction of $x_1$ to the direction of $w$ where $\theta \in [-\frac{\pi}{4},\frac{3\pi}{4}]$. And an oriented bounding box turns to a horizontal one when $\theta=0$, \eg, $\big(\mathbf x,w,h,0\big)$.}
  \label{fig:coordinates}
   \vspace{-3mm}
\end{figure}

\begin{figure*}[t!]
  \vspace{-3mm}
  \centering
  \subfigure[standard 2D convolution]{\includegraphics[width=0.2 \textwidth]{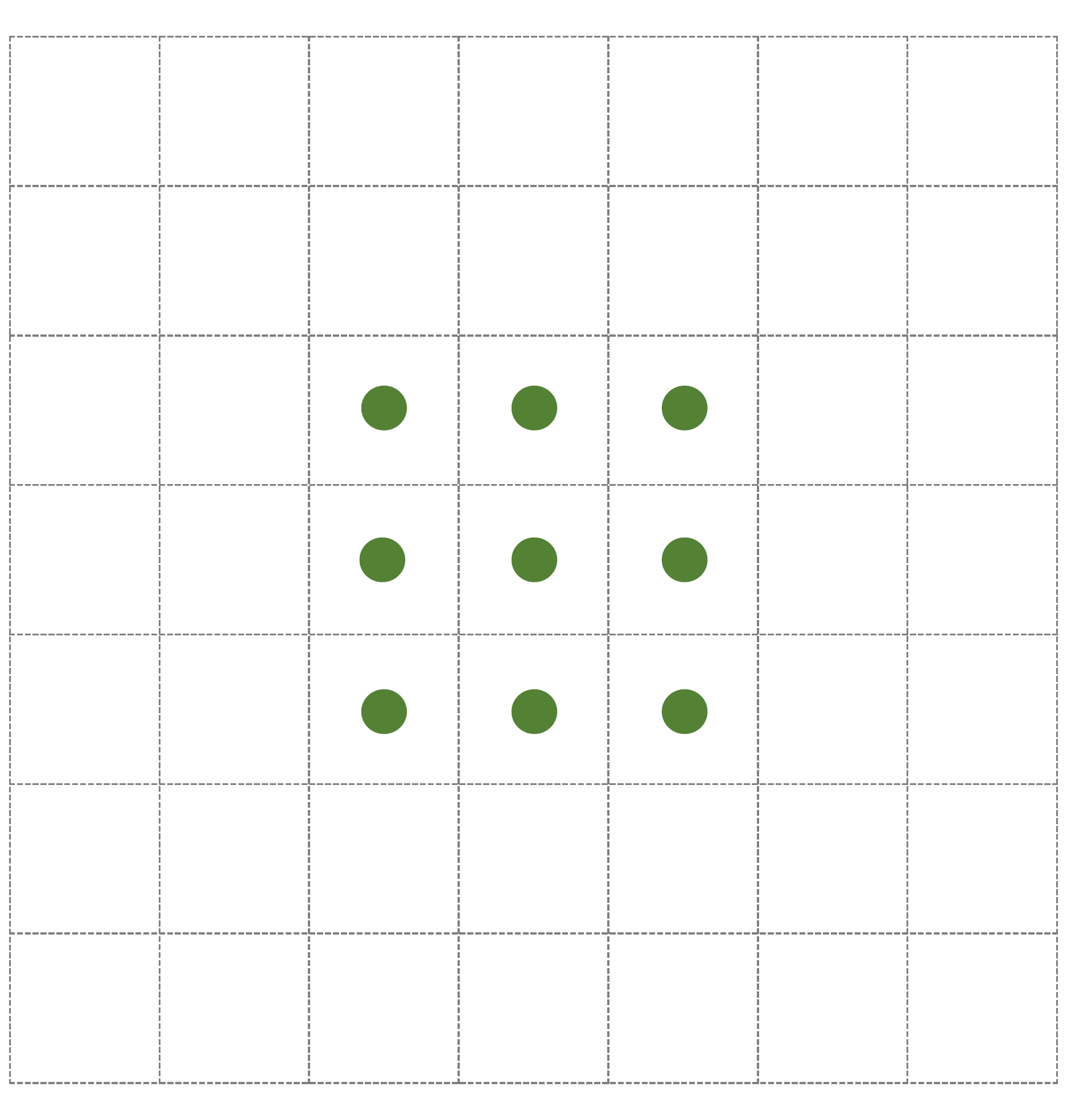}}
  \subfigure[Deformable convolution]{\includegraphics[width=0.2 \textwidth]{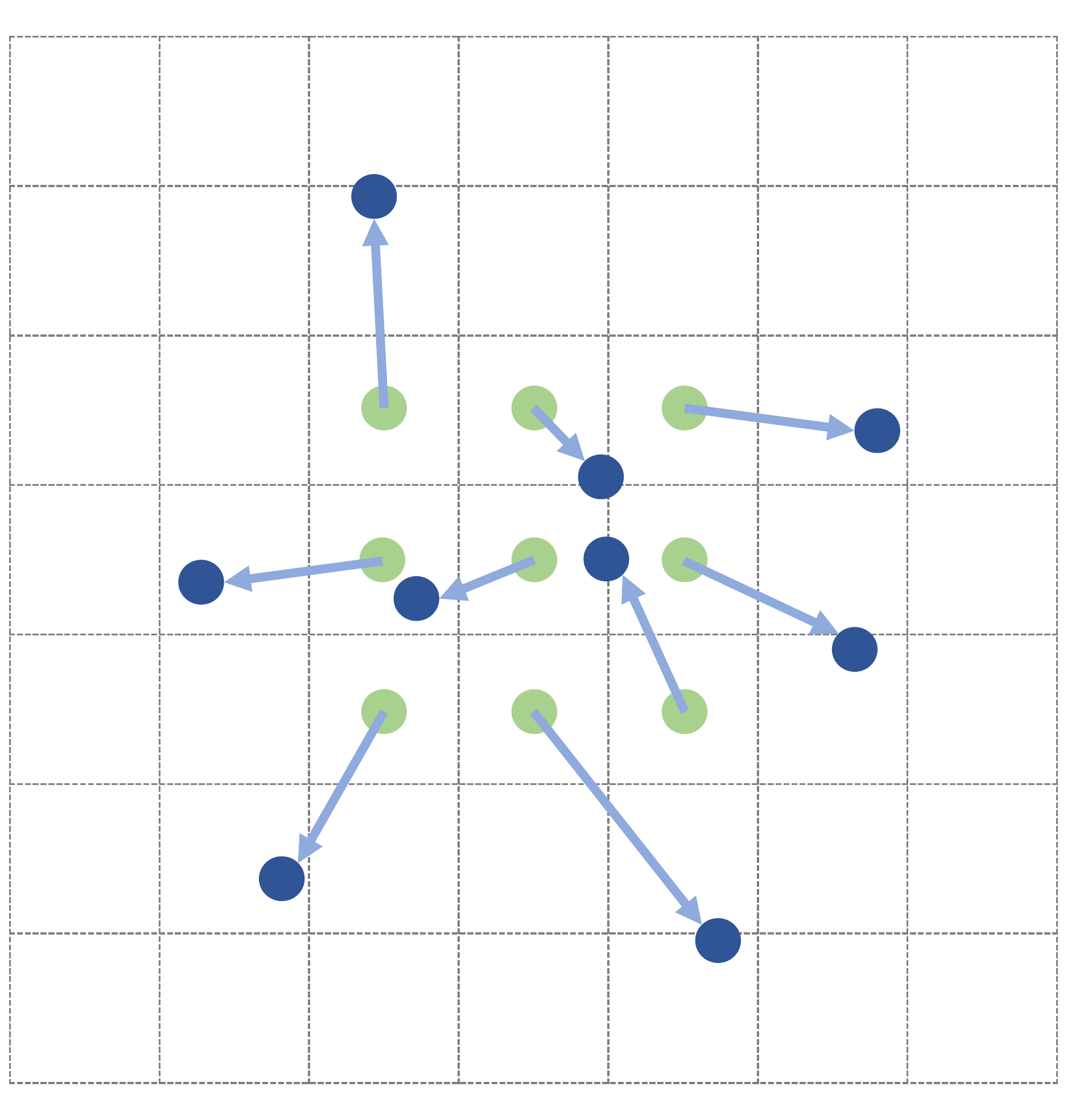}}
  \subfigure[AlignConv w. horizontal AB]{\includegraphics[width=0.2 \textwidth]{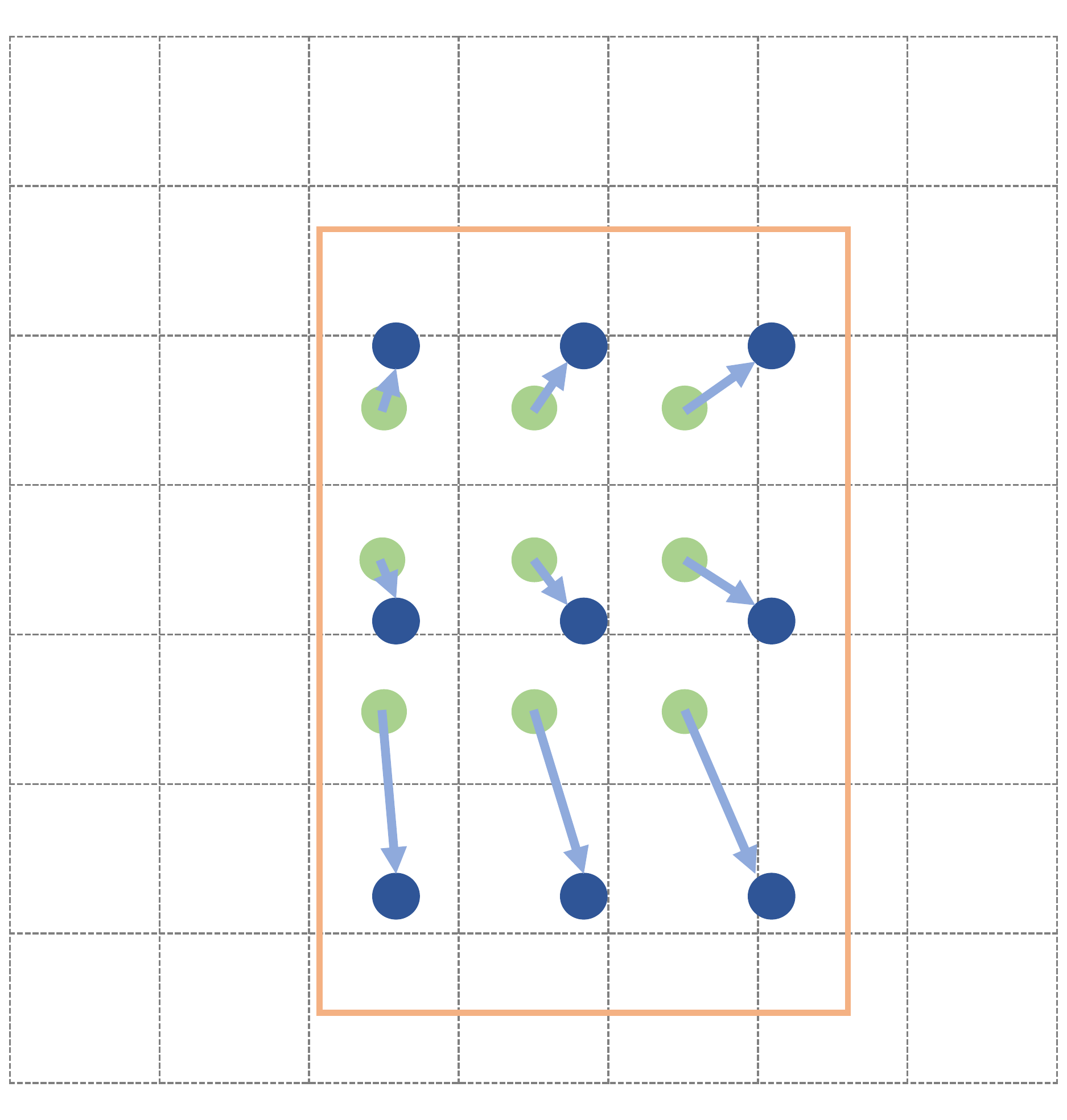}}
  \subfigure[AlignConv w. rotated AB]{\includegraphics[width=0.2 \textwidth]{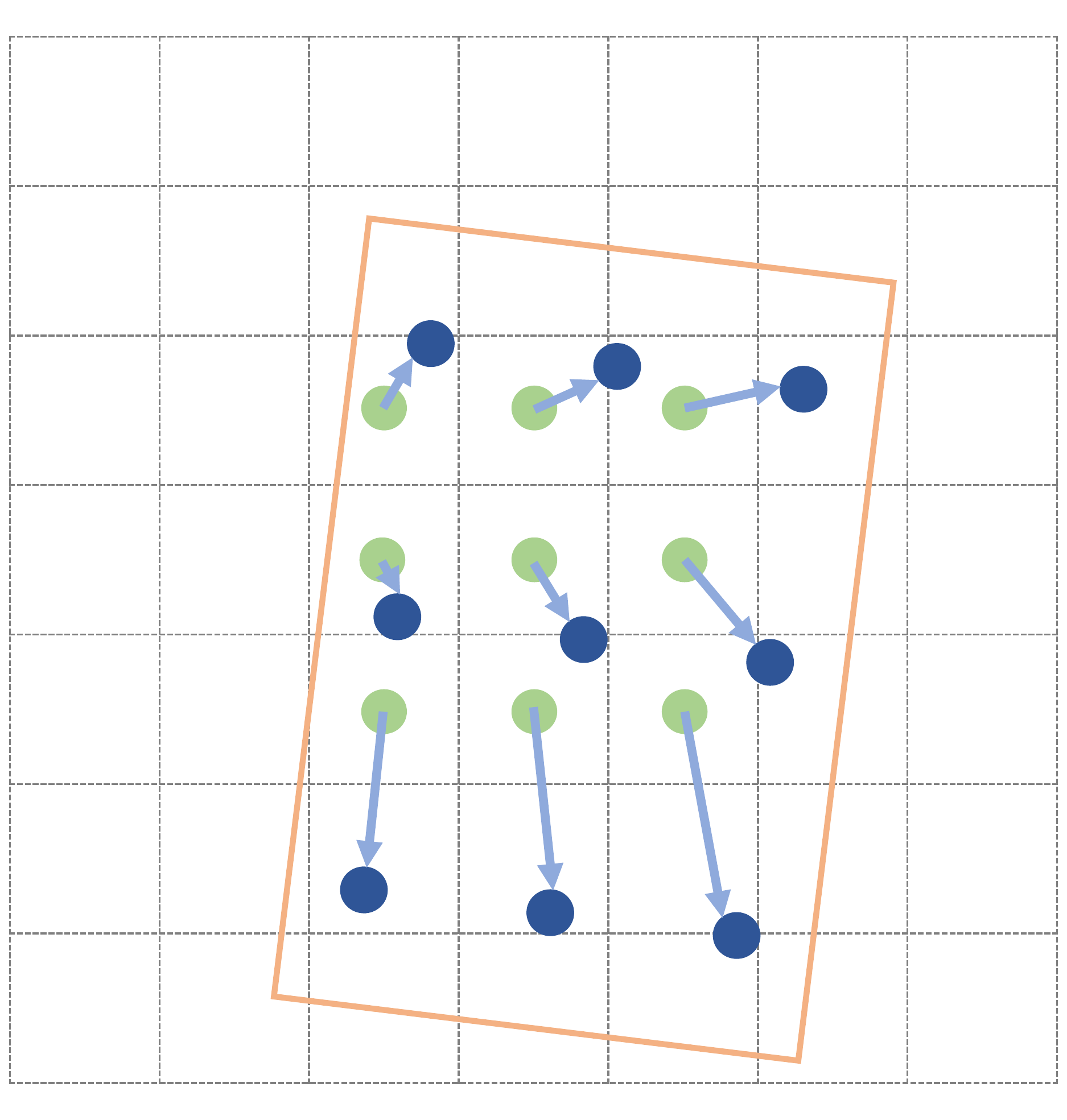}}
  \vspace{-2mm}
  \caption{Illustration of the sampling locations in different methods with 3$\times$3 kernel. (a) is the standard 2D convolution with regular sampling locations (in green dots). (b) is Deformable Convolution~\cite{dai2017dcn} with deformable sampling locations (in blue dots). (c) and (d) are two examples of our proposed AlignConv with horizontal and rotated anchor box (AB), respectively (in orange rectangle). The blue arrows mean the offset field.}
  \label{fig:diff_convs}
   \vspace{-3mm}
\end{figure*}

\section{Proposed Method}
\label{sec:methods}
In this section, we first enable RetinaNet for oriented object detection and select it as our baseline in Section~\ref{sec:retinanet}. Then, we detail the Alignment Convolution in Section~\ref{sec:alignconv}. The architectures of Feature Alignment Module and Oriented Detection Module are presented in Section~\ref{sec:fam} and Section~\ref{sec:odm}, respectively. Finally, we show details of the proposed S$^2$A-Net in both training and inference phases. The overall architecture is shown in Fig.~\ref{fig:S$^2$A-Net}, and the code is available at~\url{https://github.com/csuhan/s2anet}.

\vspace*{-3mm}
\subsection{RetinaNet as Baseline}
\label{sec:retinanet}
We choose a representative single-shot detector, RetinaNet~\cite{lin2017focal} as our baseline. It consists of a backbone network and two task-specific subnetworks. Feature pyramid network (FPN)~\cite{lin2017feature} is adopted as the backbone network to extract multi-scale features. Classification and regression subnetworks are fully convolutional networks with several (\ie, 4) stacked convolution layers. Moreover, Focal loss is proposed to address the extreme foreground-background class imbalance during training. 

Note that RetinaNet is designed for generic object detection, outputting horizontal bounding box (Fig.~\ref{fig:coordinates} (a)) represented as,
\begin{align*}
    \big\{\big( \mathbf x, w, h \big)\big\},
\end{align*}
with $\mathbf x=(x_1, x_2)$ as the center of the bounding box. 
In order to be compatible with oriented object detection, we replace the regression output of the RetinaNet with oriented bounding box (Fig.~\ref{fig:coordinates} (b)) as, 
\begin{align*}
\big\{\big( \mathbf x, w, h, \theta \big)\big\}, 
\end{align*}
where $\theta \in [-\frac{\pi}{4},\frac{3\pi}{4}]$ denotes the angle from the position direction of $x_1$ to the direction of the width $w$~\cite{ding2018transformer}. All other settings keep unchanged with original RetinaNet.

\vspace*{-3mm}
\subsection{Alignment Convolution}
\label{sec:alignconv}

In a standard 2D convolution, we first sample over the input feature map $\mathbf{X}$ defined on $\Omega = \{0, 1, \ldots, H-1\}\times \{0, 1, \ldots, W-1\}$ by a regular grid $\mathcal{R}=\{(r_x,r_y)\}$, and then sum up the sampled values weighted by $\mathbf{W}$. For example, the grid $\mathcal{R}=\{(-1,-1),(-1,0),\dots,(0,1),(1,1)\}$ represents a kernel size $3\times3$ and dilation 1. For each location $\mathbf{p} \in \Omega$ on the output feature map $\mathbf{Y}$, we have
\begin{equation}
    \mathbf{Y(p)=\sum_{r \in \mathcal{R}} W(r) \cdot X(p+r)}. \label{eq:stard_conv}
\end{equation}

Compared with standard convolution, Alignment Convolution (AlignConv) adds an additional offset field $\mathcal{O}$ for each location $\mathbf{p}$, that is
\begin{equation} \label{eq:alignconv}
    \mathbf{Y(p) = \sum_{\mathbf{r} \in \mathcal{R};\, \mathbf{o} \in \mathcal{O} } W(r) \cdot X(p+r+o)}.
\end{equation}
As shown in Fig.~\ref{fig:diff_convs} (c) and (d), for location $\mathbf{p}$, the offset field $\mathcal{O}$ is calculated as the difference between anchor-based sampling locations and regular sampling locations (\ie, $\mathbf{p+r}$). Let $(\mathbf x,w,h,\theta)$ represent the corresponding anchor box at location $\mathbf{p}$. For each $\mathbf r \in \mathcal{R}$, the anchor-based sampling location $\mathbf{L_p^r}$ can be defined as
\begin{align} \label{eq:offset_sampling_location}
    \mathbf{L_{p}^r} &= \frac{1}{S} \big( \mathbf x + \frac{1}{k}(w, h) \cdot \mathbf r \cdot R^{T}(\theta) \big),
\end{align}
where $k$ indicates the kernel size, $S$ denotes the stride of the feature map, and $R(\theta)=(\cos \theta, -\sin \theta; \, \sin \theta, \cos \theta )^T$ is the rotation matrix, respectively.
The offset field $\mathcal{O}$ at location $\mathbf{p}$ is
\begin{equation} \label{eq:offset_func}
    \mathbf{\mathcal{O} = \{ L_p^r - p - r\}_{r \in R}}.
  \end{equation}
In this way, we can transform the axis-aligned convolutional features $\mathbf{X(p)}$ of a given location $\mathbf{p}$ into arbitrary oriented ones based on the corresponding anchor box. 

{\bf Comparisons with other convolutions.} As shown in Fig.~\ref{fig:diff_convs}, standard convolution samples over the feature map by a regular grid. DeformConv learns an offset field to augment the spatial sampling locations. However, it may sample from wrong locations with weak supervision, especially for densely packed objects. Our proposed AlignConv extracts grid-distributed features with the guide of anchor boxes by adding an additional offset field. Different from DeformConv, the offset field in AlignConv is inferred from the anchor boxes directly. The examples in Fig.~\ref{fig:diff_convs} (c) and (d) illustrate that our AlignConv can extract accurate features inside the anchor boxes.

\subsection{Feature Alignment Module (FAM)}
\label{sec:fam}
This section introduces the FAM that consists of an Anchor Refinement Network and an Alignment Convolution Layer, illustrated in Fig.~\ref{fig:S$^2$A-Net} (c).

{\bf Anchor Refinement Network.} The Anchor Refinement Network (ARN) is a light network with two parallel branches: an anchor classification branch (not shown in the figure) and an anchor regression branch. The anchor classification branch classifies anchors into different categories and the anchor regression branch refines horizontal anchors into rotated anchors with high-quality. 
By default, since we only need the regressed anchor boxes to adjust the sampling locations in AlignConv, the classification branch is discarded in the inference phase to speed up the model. But for a fast version of S$^2$A-Net, for which the output of ARN is adopted to produce the final predictions (see Section~\ref{sec:det_large}), the classification branch is reserved. Following the one-to-one fashion in anchor-free detectors, we preset one squared anchor for each location in the feature map. And we do not filter out the predictions with low confidence because we notice that some negative predictions turn to positive in the final predictions. 

\begin{figure}[!t]
  \centering
  \includegraphics[width=0.62 \linewidth]{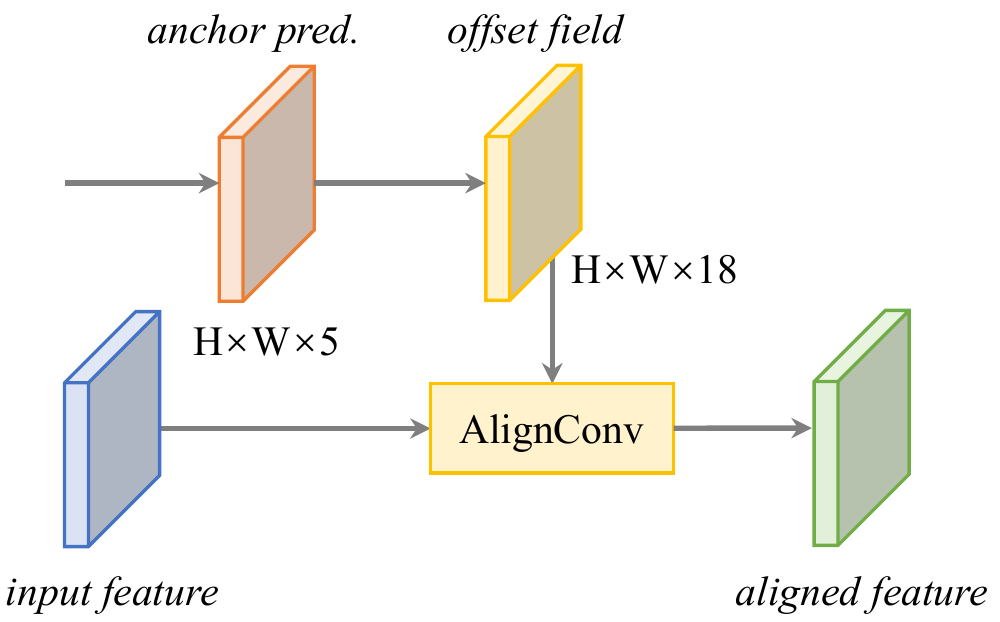}
  \vspace{-3mm}
  \caption{Alignment Convolution Layer. It takes the input feature and the anchor prediction (\emph{pred.}) map as inputs and produces aligned features. }
  \label{fig:alignconv}
   \vspace{-3mm}
\end{figure}

{\bf Alignment Convolution Layer.} With AlignConv embedded, we forms an Alignment Convolution Layer (ACL) which is shown in Fig.~\ref{fig:alignconv}. Specifically, for each location in the $H\times W\times5$ anchor prediction map, we first decode it into absolute anchor boxes $(\mathbf x, w,h,\theta)$. Then the offset field calculated by Eq.~\eqref{eq:offset_func} along with the input feature are fed into AlignConv to extract aligned features. 
Note for each anchor box (5-dimension), we regularly sample 9 (3 rows and 3 columns) points to obtain the 18-dimension offset field (\ie, the \emph{x}-offset and \emph{y}-offset of 9 points, see the blue arrows in Fig.~\ref{fig:diff_convs} (c) and (d)). Besides, it should be emphasized that ACL is a light convolution layer with negligible speed latency in offset field calculation.

\subsection{Oriented Detection Module (ODM)}
\label{sec:odm}
As shown in Fig.~\ref{fig:S$^2$A-Net} (d), the Oriented Detection Module (ODM) is proposed to alleviate the inconsistency between classification score and localization accuracy and then performs accurate object detection. We first adopt active rotating filters (ARF)~\cite{zhou2017orn} to encode the orientation information. An ARF is a $k\times k\times N$ filter that actively rotates $N-1$ times during convolution to produce a feature map with $N$ orientation channels ($N$ is 8 by default). For a feature map $\mathbf{X}$ and an ARF $\mathbf{F}$, the $i$-th orientation output of $\mathbf{Y}$ can be denoted as
\begin{align} \label{eq:orconv}
  \mathbf{Y}^{(i)}=\sum_{n=0}^{N-1} \mathbf{F}_{\theta_i}^{(n)} \cdot \mathbf{X}^{(n)}, \, \theta_i=i \frac{2 \pi}{N}, \,i=0, \dots, N-1,
\end{align}
where $\mathbf{F}_{\theta_i}$ is the clockwise $\theta_i$-rotated version of $\mathbf{F}$, $\mathbf{F}_{\theta_i}^{(n)}$ and $\mathbf{X}^{(n)}$ are the $n$-th orientation channel of $\mathbf{F}_{\theta_i}$ and $\mathbf{X}$, respectively. Applying ARF to the convolution layer, we can obtain orientation-sensitive features with explicitly encoded orientation informations. The bounding box regression task benefits from the orientation-sensitive features, while the object classification task requires invariant features. Following~\cite{zhou2017orn}, we aims to extract orientation-invariant features by pooling the orientation-sensitive features. This is simply done by choosing the orientation channel with strongest response as the output feature $\mathbf{\hat{X}}$.
\begin{align} \label{eq:orpool}
 \mathbf{\hat{X}} = \max\,\mathbf{X}^{(n)}, \, 0<n<N-1. 
\end{align}
In this way, we can align the feature of objects with different orientations, toward robust object classification. Compared with the orientation-sensitive feature, the orientation-invariant feature is efficient with fewer parameters. For example, an $H\times W \times 256$ feature map with 8 orientation channels becomes $H \times W\times 32$ after pooling. Finally, we feed the orientation-sensitive feature and orientation-invariant feature into two subnetworks to regress the bounding boxes and classify the categories, respectively.

\subsection{Single-Shot Alignment Network}
\label{sec:san}
We adopt RetinaNet as the baseline, including its network architecture and most parameter settings, and form S$^2$A-Net based on the combination of FAM and ODM. In the following, we detail S$^2$A-Net in both training and inference phases.

{\bf Regression targets.} Following previous works, we give the parameterized regression targets as:
\begin{equation}
    \begin{aligned}
    \Delta \mathbf{x}_g &=   \big(\mathbf{x}_g - \mathbf{x} \big)R(\theta) \cdot (\frac{1}{w},\frac{1}{h}), \\
    (\Delta w_g, \Delta h_g) &= \log (w_g, h_g) - \log (w, h), \\
    \Delta{\theta}_g &= \frac{1}{\pi}(\theta_g-\theta+k\pi),
\end{aligned}
\label{eq:targets}
\end{equation}
where $\mathbf x_g$, $\mathbf x$ are for the ground-truth box and the anchor box respectively (likewise for $w, h, \theta$). And $k$ is an integer to ensure $(\theta_g-\theta+k\pi) \in [-\frac{\pi}{4},\frac{3\pi}{4}]$ (see Fig.~\ref{fig:coordinates}). In FAM, we set $\theta=0$ to represent a horizontal anchor. Then the regression targets can be expressed by Eq.~\eqref{eq:targets}. In ODM, we first decode the output of FAM and then re-compute the regression targets by Eq.~\eqref{eq:targets}. 

{\bf Matching strategy.} We adopt IoU as the metrics, and an anchor box can be assigned to positive (or negative) if its IoU is greater than a foreground threshold (or less than a background threshold, respectively). Different from the IoU between horizontal bounding boxes, we calculate the IoU between two oriented bounding boxes. By default, we set the foreground threshold as $0.5$ and the background threshold as $0.4$ in both FAM and ODM.

{\bf Loss function.} The loss of S$^2$A-Net is a multi-task one which consists of two parts, \ie, the loss of FAM and the loss of ODM. For each part, we assign a class label to each anchor/refined anchor and regress its location. The loss function can be defined as:
\begin{equation}
    \begin{aligned}
    \mathcal{L}= & \frac{1}{N_{F}} \big( \sum_i \mathcal{L}_c(c_i^F, l_i^*) + \sum_i \mathbf{1}_{[l_i^* \geq 1]} \mathcal{L}_r(\mathbf x_i^F, \mathbf g_i^*) \big) \\
                 + &\frac{\lambda}{N_{O}} \big( \sum_i \mathcal{L}_c(c_i^O,l_i^*) + \sum_i \mathbf{1}_{[l_i^* \geq 1]}\mathcal{L}_r(\mathbf x_i^O, \mathbf g_i^*) \big),
\end{aligned}
\label{eq:loss}
\end{equation}
where $\lambda$ is a loss balance parameter, $\mathbf{1}_{[\cdot]}$ is an indicator function, $N_{F}$ and $N_{O}$ are the numbers of positive samples in the FAM and ODM respectively, $i$ is the index of a sample in a mini-batch. $c_i^F$ and $\mathbf x_i^F$ are the predicted category and refined locations of the anchor $i$ in FAM. $c_i^O$ and $\mathbf x_i^O$ are the predicted object category and locations of the bounding box in ODM. $l_i^*$ and $\mathbf g_i^*$ are the ground-truth category and locations of the anchor $i$. The Focal loss~\cite{lin2017focal} and smooth $L1$ loss are adopted as the classification loss $\mathcal{L}_c$ and the regression loss $\mathcal{L}_r$, respectively.

{\bf Inference.} S$^2$A-Net is a fully convolutional network and we can simply forward an image through the network without complex RoI operation. Specifically, we pass the input image to the backbone network to extract pyramid features. Then the pyramid features are fed into FAM to produce refined anchors and aligned features. After that, ODM encodes the orientation information to produce the predictions with high confidence. Finally, we choose top-$k$ (\ie, 2000) predictions and adopt NMS to yield the final detections.

\section{Experiments and Analysis}
\label{sec:experiments}
\subsection{Datasets}
\label{sec:dataset}
{\bf DOTA~\cite{xia2018dota}.} It is a large aerial image dataset for oriented objects detection which contains 2806 images with the size ranges from $800\times800$ to $4000\times4000$ and 188282 instances of 15 common object categories includes: Plane (PL), Baseball diamond (BD), Bridge (BR), Ground track field (GTF), Small vehicle (SV), Large vehicle (LV), Ship (SH), Tennis court (TC), Basketball court (BC), Storage tank (ST), Soccer-ball field (SBF), Roundabout (RA), Harbor (HA), Swimming pool (SP), and Helicopter (HC).

Both training and validation sets are used for training, and the testing set is used for testing. Following~\cite{xia2018dota}, we crop a series of $1024\times1024$ patches from original images with a stride of 824. We only adopt random horizontal flipping during training to avoid over-fitting and no other tricks are utilized if not specified. For fair comparison with other methods, we adopt data augmentation (\ie, random rotation) in the training phase. For multi-scale experiments, we firstly resize original images at three scales (0.5, 1.0 and 1.5) and then crop them into 1024$\times$1024 patches with a stride of 512.

{\bf HRSC2016~\cite{liu2017hrsc2016}.} It is a high resolution ship recognition dataset annotated with oriented bounding boxes which contains 1061 images, and the image size ranges from $300\times300$ to $1500\times900$. We use the training (436 images) and validation (181 images) sets for training and the testing set (444 images) for testing. All images are resized to $(800, 512)$ without changing the aspect ratio. Horizontal flipping is applied during training.

\subsection{Implementation Details}
\label{sec:impl_detail}
We adopt ResNet101 FPN as the backbone network for fair comparison with other methods, and ResNet50 FPN is adopted for other experiments if not specified. For each level of pyramid features (\ie, $P_3$ to $P_7$), we preset one squared anchor per location with a scale of 4 times the total stride size (\ie, 32, 64, 128, 256, 512). The loss balance parameter $\lambda$ is set to 1. The hyperparameters of Focal loss $\mathcal{L}_c$ are set to $\alpha=0.25$ and $\gamma=2.0$. We adopt the same training schedules as mmdetection~\cite{chen2019mmdetection}. We train all models in 12 epochs for DOTA and 36 epochs for HRSC2016. SGD optimizer is adopted with an initial learning rate of 0.01 and the learning rate is divided by 10 at each decay step. The momentum and weight decay are 0.9 and 0.0001, respectively. We adopt learning rate warmup for 500 iterations. We use 4 V100 GPUs with a total batch size of 8 for training and a single V100 GPU for inference by default. The time of post-processing (\eg, NMS) is included in all experiments.

\subsection{Ablation Studies}
In this section, we conduct a series of experiments on the testing set of DOTA to validate the effectiveness of our method. ResNet50 FPN is adopted as the backbone in all experiments. Note that we extend the \texttt{flops\_counter} tool in mmdetection~\cite{chen2019mmdetection} to calculate the FLOPs of our method.

\begin{table}[!t]
  \caption{Results of different RetinaNet on DOTA. Depth indicates the number of convolution layer in two subnetworks of RetinaNet.}
  \vspace{-3mm}  
\begin{center}
    \begin{tabular}{c|c|c|c|c|c|c} \hline
       & Model        &\#Anchor & Depth & mAP    &GFLOPs & Param   \\ \hline
    (a)& RetinaNet    &  9      & 4     & 68.05  &215.92 & 36.42 M \\
    (b)& RetinaNet    &  9      & 2     & 67.64  &164.38 & 34.06 M \\
    (c)& RetinaNet    &  1      & 2     & 67.00  &156.33 & 33.69 M \\ \hline
    \end{tabular}
\end{center}
    \label{tab:retinenet}
    \vspace{-3mm}  
\end{table}

\begin{table*}[!t]
  \caption{Comparing Alignment Convolution (AlignConv) with other convolution methods. We compare our AlignConv with the standard convolution (Conv), Deformable Convolution (DeformConv) and Guided Anchoring Deformable Convolution (GA-DeformConv).}
  \vspace{-3mm}
  \centering
\resizebox{\textwidth}{!}{%
    \begin{tabular}{c|ccccccccccccccc|c|c} \hline
    Methods       & PL & BD    & BR & GTF   & SV    & LV    & SH  & TC    & BC    & ST    & SBF   & RA    & HA & SP    & HC    & mAP    & GFLOPs  \\ \hline
    Conv          & 88.87 & 76.34 & 46.42  & 67.53 & 77.21 & 74.80 & 82.27 & 90.79 & 81.22 & 85.02 & 50.99 & 61.10 & 63.54  & 67.24 & 53.25 & 71.11  & \textbf{196.62} \\
    DeformConv    & 88.96 & 80.23 & 45.92  & 67.51 & 77.10 & 74.23 & 84.28 & 90.81 & 81.47 & 85.56 & 54.19 & \textbf{64.11} & 64.85  & 68.13 & 48.34 & 71.71  & 198.02 \\
    GA-DeformConv & 88.72 & 79.56 & 46.19  & 65.41 & 76.86 & 74.96 & 79.44 & 90.78 & 80.99 & 84.73 & 55.31 & 63.17 & 62.07  & 67.69 & 54.12 & 71.33   & 197.92 \\
    AlignConv     & \textbf{89.11} & \textbf{82.84} & \textbf{48.37}  & \textbf{71.11} & \textbf{78.11} & \textbf{78.39} & \textbf{87.25} & \textbf{90.83} & \textbf{84.90} & \textbf{85.64} & \textbf{60.36} & 62.60 & \textbf{65.26}  & \textbf{69.13} & \textbf{57.94} & \textbf{74.12} & 198.03 \\ \hline
    \end{tabular}%
}
  \label{tab:conv_type}
  \vspace{-3mm}
\end{table*}

\begin{figure}[t!]
  \centering
  \includegraphics[width=0.93\linewidth]{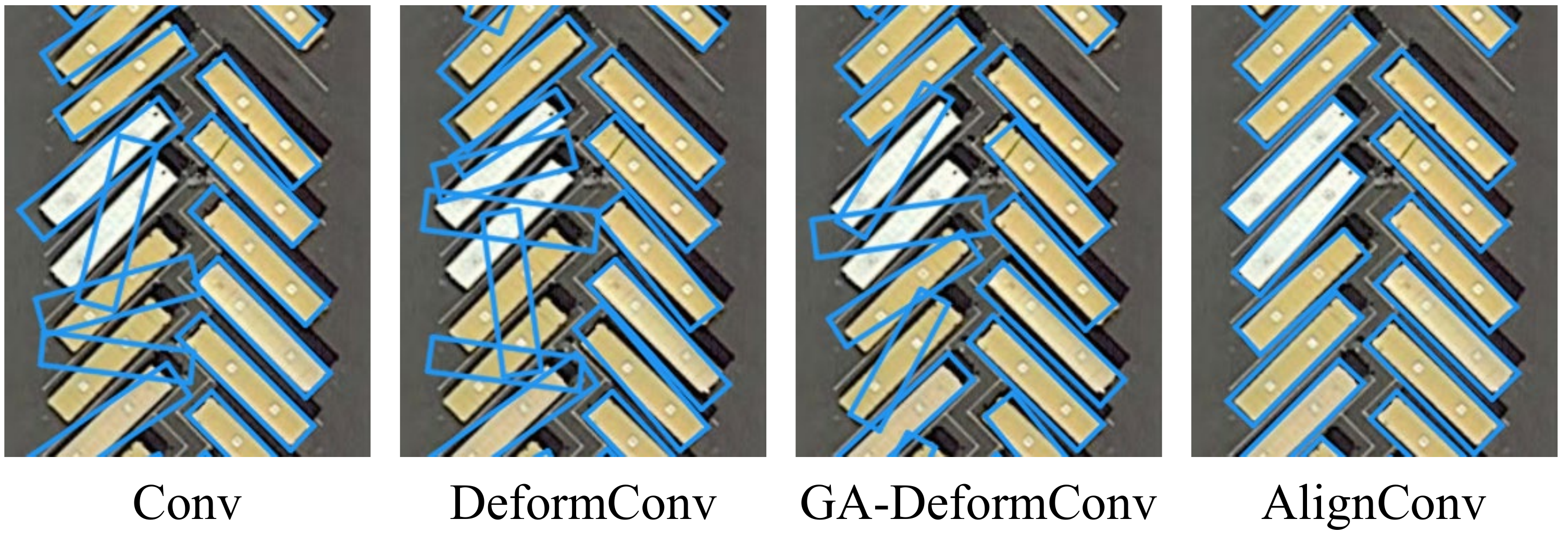}
  \vspace{-3mm}
  \caption{Qualitative comparison of different convolution methods. The blue bounding box indicates the prediction of large vehicle.}
  \label{fig:conv_compare}
  \vspace{-3mm}
\end{figure}

\begin{table}[t!]
  \caption{Ablation studies. We choose a light RetinaNet (shown in Table~\ref{tab:retinenet} (c)) as the baseline, and experiment different settings of S$^2$A-Net, \ie, Anchor Refinement Network (ARN), Alignment Convolution Layer (ACL) and active rotating filters (ARF). }
  \vspace{-3mm}
  \centering
\resizebox{\linewidth}{!}{%
\begin{tabular}{c|c|cccccc} \hline
    & Baseline  & \multicolumn{6}{c}{Different Settings of S$^2$A-Net}  \\ \hline
ARN &           &          &\checkmark&          &\checkmark&\checkmark&\checkmark \\
ACL &           &          &          &          &          &\checkmark&\checkmark \\
ARF &           &          &          &\checkmark&\checkmark&          &\checkmark \\ \hline
mAP & 67.00     &  68.26   &   71.17  &   68.35  & 71.11    &  73.24   &74.12 \\ \hline
\end{tabular} 
}
  \label{tab:acl_arf}
  \vspace{-3mm}
\end{table}

\begin{table}[t!]
  \caption{Experiments of different network designs. We explore the network design in FAM and ODM with different number of layers. Setting (d) is the default setting of our proposed method shown in Fig.~\ref{fig:S$^2$A-Net}.}
  \vspace{-3mm}

\begin{center}
    \begin{tabular}{c|c|c|c|c|c|c} \hline
        & Model            &FAM  &ODM    & mAP          & GFLOPs          & Param \\ \hline
    (a) & RetinaNet        & -   & -     &68.05         & 215.92          & 36.42 M      \\
    (b) & S$^2$A-Net       & 1   & 1     &73.04         & \textbf{159.27} & \textbf{33.25 M} \\
    (c) & S$^2$A-Net       & 1   & 3     &72.89         & 210.81          & 35.61 M      \\
    (d) & S$^2$A-Net       & 2   & 2     &\textbf{74.12}& 198.03          & 35.02 M      \\
    (e) & S$^2$A-Net       & 1   & 3     &72.86         & 185.04          & 34.43 M      \\
    (f) & S$^2$A-Net       & 4   & 4     &73.30         & 275.22          & 38.57 M      \\ \hline
    \end{tabular}
    \end{center}
    \label{tab:net_design}
  \vspace{-3mm}
\end{table}

{\bf RetinaNet as baseline.} As a single-shot detector, RetinaNet is fast enough. However, any module added to it will introduce more computations. We experiment different architectures and settings on RetinaNet. As shown in Table~\ref{tab:retinenet} (a), RetinaNet achieves a mAP of 68.05\% with 215.92 GFLOPs and 36.42 M parameters, indicating that our baseline is solid. If the depth of RetinaNet head changes from 4 to 2, the mAP drops 0.41\% and the FLOPs (\emph{resp.} parameters) reduce 51.54 G (\emph{resp.} 2.36 M). Furthermore, if we set one anchor per location (Table~\ref{tab:retinenet} (c)), the FLOPs reduces 28\% with a accuracy drop of 1.5\% compared with Table~\ref{tab:retinenet} (a). The results show that a light detection head and few anchors can also achieve competitive performance and better speed-accuracy trade-off.

{\bf Effectiveness of AlignConv.} As discussed in Section~\ref{sec:alignconv}, we compare AlignConv with other methods to validate its effectiveness. We only replace AlignConv with other convolution methods and keep other settings unchanged. Besides, we also add comparison with Guided Anchoring DeformConv (GA-DeformConv)~\cite{wang2019region}. Note that the offset field of GA-DeformConv is learned from the anchor prediction map in ARN by a $1\times1$ convolution. 

As shown in Table~\ref{tab:conv_type}, AlignConv surpasses other methods by a big margin. Compared with the standard convolution, AlignConv improves about 3\% mAP while only introduces 1.41 GFLOPs computation. Besides, AlignConv improves the performance for almost all categories, especially for those categories with large aspect ratios (\eg, bridge), densely distribution (\eg, small vehicles and large vehicles) and fewer instances (\eg, helicopters). On the contrary, DeformConv and GA-DeformConv only achieve 71.71\% and 71.33\% mAP, respectively. The qualitative comparison in Fig.~\ref{fig:conv_compare} shows that AlignConv achieves accurate bounding box regression in detecting densely packed and arbitrary oriented objects, while other methods with implicit learning get poor performance.

{\bf Effectiveness of ARN and ARF.} To evaluate the effectiveness of ARN and ARF, we experiment different settings of S$^2$A-Net. If ARN is discarded, then FAM and ODM share the same initial anchors without refinement. If ARF is discarded, we replace the ARF layer with the standard convolution layer. As shown in Table~\ref{tab:acl_arf}, without ARN, ACL and ARF, our method achieves 68.26\% mAP, about 1.26\% mAP higher than the baseline method. This is mainly because we add supervisions in both FAM and ODM. With the participation of ARN, we obtain 71.17\% mAP, showing that anchor refinement is important to the final predictions in ODM.

Besides, we find ARF dose nothing for performance improvement without the participation of ACL, \ie, applying ARF or the combination of ARN and ARF to our method only achieve 68.35\% and 71.11\% mAP, respectively.
However, if we put ACL and ARF together, there is an obvious improvement, from 73.24\% to 74.12\%. We argue that CNNs are not rotation-invariant, and even we can extract accurate features to represent the object, the corresponding features are still rotation-sensitive. So the participation of ARF augments the orientation information explicitly, leading to better regression and classification.

{\bf Network design.} As shown in Table~\ref{tab:net_design}, we explore different network designs in FAM and ODM. Compared with the baseline method in Table~\ref{tab:net_design} (a), we can conclude that S$^2$A-Net is not only an effective detector with high detection accuracy, but also an efficient detector in both speed and parameters. 
The results in Table~\ref{tab:net_design} (b)-(f) show that our proposed method is insensitive to the depth of the network and the performance improvements mainly come from our novel alignment mechanism. Besides, as the number of layers increases, there is a performance drop from Table~\ref{tab:net_design} (d) to (f). We hypothesize that deeper networks with a larger receptive field may hinder the detection performance of small size objects. Moreover, the setting (d), for which the number of layers in FAM and ODM is the same, obtains the highest mAP among (c)-(e), while (c) and (e) have a significant drop in mAP, showing that similar receptive field in FAM and ODM is more balancing for high quality object detection.

\begin{table}[!t]
  \caption{Comparison of different settings detecting on large images in DOTA. {\bf Stride} is the cropping stride referred in Section~\ref{sec:dataset}. {\bf \#Image} means the number of images or chips. {\bf Output} indicates the module (\ie, FAM or ODM) used for testing. We show the inference time required for entire dataset using \emph{FP32}/\emph{FP16} with 4 V100 GPUs.}
    \vspace{-4mm}
    \begin{center}
    \begin{tabular}{c|c|c|c|c|c} \hline
    Input Size         & Stride &\#Image& Output & mAP & Time (s)   \\ \hline
    $1024 \times 1024$ & 1024   & 8143  & ODM    &71.20& 150 / 126  \\
    $1024 \times 1024$ & 824    & 10833 & ODM    &74.12& 246 / 160  \\
    $1024 \times 1024$ & 512    & 20012 & ODM    &74.62& 352 / 308  \\
    Original           & -      & 937   & ODM    &74.01& 120 / 103  \\
    Original           & -      & 937   & FAM    &70.85& 104 / 97   \\ \hline
    \end{tabular}
    \end{center}
    \label{tab:det_large}
    \vspace{-3mm}
\end{table}

\begin{table}[!t]
  \caption{Comparing S$^2$A-Net with ClusDet~\cite{Yang2019clusdet} on DOTA validation set. Following~\cite{Yang2019clusdet}, we report the accuracy of five categories (\ie, PL, SV, LV, SH and HC) with different IoU thresholds (\ie, mAP$_{.5}$, mAP$_{.75}$ and mAP$_{.5-.95}$). The results of RetinaNet and S$^2$A-Net are calculated from the axis-aligned bounding boxes of the output. \#Image means the number of images or chips. $^\dag$ indicates that the output of FAM is adopted for the final results.}
  \vspace{-2mm}
  \centering
\begin{tabular}{c|c|ccc}
\hline
Methods                        & \#Image         & mAP$_{.5-.95}$ & mAP$_{.5}$ & mAP$_{.75}$ \\ \hline
ClusDet~\cite{Yang2019clusdet} & 1055            & 32.2      & 47.6   & 39.2    \\
RetinaNet                      & \textbf{458}    & 41.6     & 70.5  & 44.2 \\
S$^2$A-Net$^\dag$ (Ours)       & \textbf{458}    & 42.7      & 72.7  & 45.3   \\
S$^2$A-Net (Ours)              & \textbf{458}    & \textbf{43.9}      & \textbf{75.8}  & \textbf{46.3}   \\ \hline
\end{tabular}
  \label{tab:compare_clusdet}
  \vspace{-3mm}
\end{table}

\begin{figure}[!t]
  \centering
  \includegraphics[width= .95\linewidth]{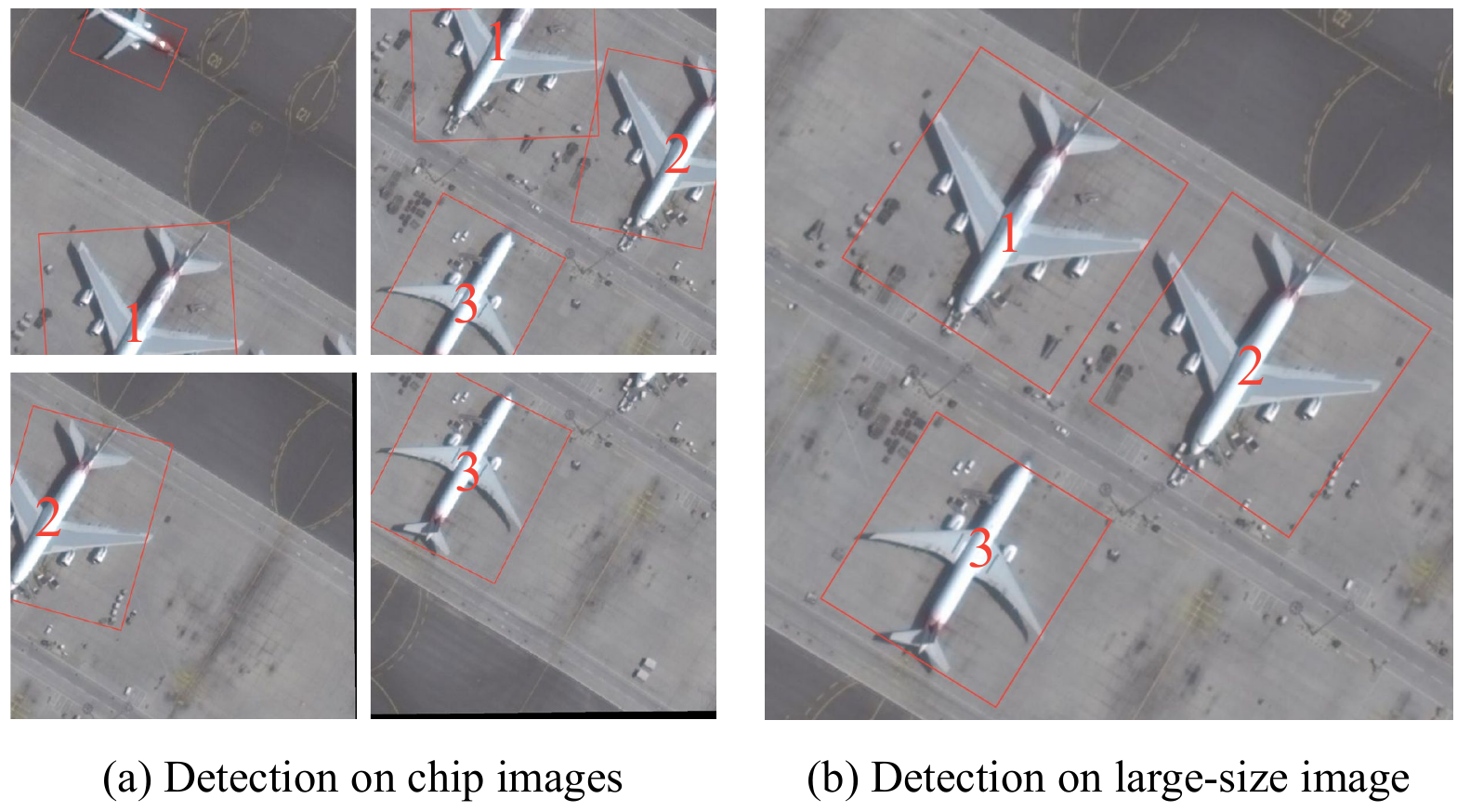}
  \vspace{-3mm}
  \caption{Qualitative comparison of detection results. We crop a large-size image into $1024\times1024$ chip images with a stride of 824. The large-size image and chip images are fed into the same network to produce detection results (\eg, planes in red boxes) without resizing. Instances with the same number are corresponding.}
  \label{fig:large_compare}
  \vspace{-4mm}
\end{figure}

\subsection{Detecting on large-size images}
\label{sec:det_large}
The size of aerial image often ranges from thousands to tens of thousands, which means more computations and memory footprint. Many previous works~\cite{xia2018dota, ding2018transformer} adopt a detection on chips strategy to alleviate this challenge, even if a chip does not contain any object. ClusDet~\cite{Yang2019clusdet} tries to address this issue by generating clustered chips, while introducing more complex operations (\eg, chip generation and results merge) and significant performance drop. As our proposed S$^2$A-Net is efficient and the architecture is flexible, we aims to detect objects on large-size images directly. 

We first explore different settings of the input size and cropping stride, and report the mAP and overall time during inference (Table~\ref{tab:det_large}). We first crop the images into $1024\times1024$ chips, and the mAP improves from 71.20\% to 74.62\% when the stride decreases from 1024 to 512. However, the number of chip images increases from 8143 to 20012, and the overall inference time increases about 135\%. If we detect on the original large-size images without cropping, the inference time has reduced by 50\% with negligible loss of accuracy. We argue that the cropping strategy makes it hard to detect objects around the boundary (Fig.~\ref{fig:large_compare}). Besides, if we adopt the output of FAM for detection and Floating-Point 16 (FP16) to speed up the inference, we can reduce the inference time to 97 seconds with a mAP of 70.85\%. 
Compared our S$^2$A-Net with ClusDet~\cite{Yang2019clusdet} (Table~\ref{tab:compare_clusdet}), our method only process 458 images and outperforms ClusDet by a large margin. If we adopt the output of FAM for evaluation, we still achieve 42.7\% mAP$_{.5-.95}$ and 72.7\% mAP$_{.5}$. The result demonstrates that out method is efficient and effective, and our detection strategy can achieve better speed-accuracy trade-off.

\begin{table*}[t!]
  \caption{Comparisons with state-of-the-art methods on DOTA. R-101-FPN stands for ResNet 101 with FPN (likewise R-50-FPN), and H-104 stands for Hourglass 104. $^\dag$ indicates training and testing without data augmentation. $^{\ddagger}$ denotes the input is the original images other than chip images. $^*$ means multi-scale training and testing.}
  \vspace{-4mm}
   \begin{center}
    \resizebox{\textwidth}{!}{%
    \begin{tabular}{c|c|ccccccccccccccc|c|c} \hline
    Methods           & Backbone &PL& BD & BR & GTF & SV & LV & SH & TC & BC & ST & SBF & RA & HA & SP & HC & mAP & FPS \\ \hline
    \emph{two-stage:}  &     &    &       &     &    &    &      &    &    &    &     &    &        &    &    &     &     \\ 
    FR-O~\cite{xia2018dota}                    & R-101 &79.42&77.13&17.70&64.05&35.30&38.02&37.16&89.41&69.64&59.28&50.30&52.91&47.89&47.40&46.30&54.13& -    \\ 
    Azimi \etal~\cite{azimi2018towards}        & R-101-FPN &81.36&74.30&47.70&70.32&64.89&67.82&69.98&90.76&79.06&78.20&53.64&62.90&67.02&64.17&50.23&68.16& -    \\ 
    RoI Trans.$^*$~\cite{ding2018transformer}  & R-101-FPN &88.64&78.52&43.44&75.92&68.81&73.68&83.59&90.74&77.27&81.46&58.39&53.54&62.83&58.93&47.67&69.56& 5.9  \\ 
    CADNet~\cite{zhang2019cad}                 & R-101-FPN &87.80&82.40&49.40&73.50&71.10&63.50&76.60&\textbf{90.90}&79.20&73.30&48.40&60.90&62.00&67.00&62.20&69.90& -    \\ 
    SCRDet~\cite{yang2019scrdet}               & R-101-FPN   &\textbf{89.98}&80.65&52.09&68.36&68.36&60.32&72.41&90.85&\textbf{87.94}&86.86&65.02&66.68&66.25&68.24&65.21&72.61& -    \\ 
    Xu \etal~\cite{xu2019gliding}              & R-101-FPN &89.64&\textbf{85.00}&52.26&77.34&73.01&73.14&86.82&90.74&79.02&86.81&59.55&\textbf{70.91}&72.94&70.86&57.32&75.02& 10.0    \\ 
    CenterMap-Net~\cite{wang2020centermap}     & R-50-FPN &88.88&81.24&53.15&60.65&78.62&66.55&78.10&88.83&77.80&83.61&49.36&66.19&72.10&72.36&58.70&71.74& - \\
    CenterMap-Net$^*$~\cite{wang2020centermap} &R-101-FPN &89.83&84.41&54.60&70.25&77.66&78.32&87.19&90.66&84.89&85.27&56.46&69.23&74.13&71.56&66.06&76.03&- \\ \hline
    \emph{one-stage}:  &       &    &       &     &    &    &      &    &    &    &     &    &        &    &    &     &     \\ 
    RetinaNet~\cite{lin2017focal}              & R-101-FPN &88.82&81.74&44.44&65.72&67.11&55.82&72.77&90.55&82.83&76.30&54.19&63.64&63.71&69.73&53.37&68.72& 12.7    \\ 
    DRN~\cite{pan2020dynamic}                  & H-104     &88.91&80.22&43.52&63.35&73.48&70.69&84.94&90.14&83.85&84.11&50.12&58.41&67.62&68.60&52.50&70.70& - \\ 
    DRN$^*$~\cite{pan2020dynamic}                  & H-104     &89.71&82.34&47.22&64.10&76.22&74.43&85.84&90.57&86.18&84.89&57.65&61.93&69.30&69.63&58.48&73.23& - \\
    R$^3$Det~\cite{yang2019r3det}              & R-101-FPN &89.54&81.99&48.46&62.52&70.48&74.29&77.54&90.80&81.39&83.54&61.97&59.82&65.44&67.46&60.05&71.69& -    \\ 
    R$^3$Det~\cite{yang2019r3det}              & R-152-FPN &89.49&81.17&50.53&66.10&70.92&78.66&78.21&90.81&85.26&84.23&61.81&63.77&68.16&69.83&67.17&73.74& - \\
    S$^2$A-Net$^{\dag}$ (Ours)                 & R-50-FPN &89.11&82.84&48.37&71.11&78.11&78.39&87.25&90.83&84.90&85.64&60.36&62.60&65.26&69.13&57.94&74.12& 16.0    \\ 
    S$^2$A-Net$^{\dag \ddagger}$ (Ours)            & R-50-FPN  &89.11&81.51&48.75&72.85&78.23&76.77&86.95&90.84&83.59&85.52&62.70&61.63&66.55&68.94&56.24&74.01&\textbf{22.6} \\
    S$^2$A-Net (Ours)                          & R-101-FPN &88.70&81.41&54.28&69.75&78.04&80.54&88.04&90.69&84.75&86.22&65.03&65.81&76.16&73.37&58.86&76.11& 12.7    \\ 
    S$^2$A-Net$^*$ (Ours)                      & R-50-FPN  &88.89&83.60&\textbf{57.74}&\textbf{81.95}&79.94&\textbf{83.19}&89.11&90.78&84.87&\textbf{87.81}&70.30&68.25&78.30&77.01&\textbf{69.58}&\textbf{79.42}&16.0     \\
    S$^2$A-Net$^*$ (Ours)                      & R-101-FPN &89.28&84.11&56.95&79.21&\textbf{80.18}&82.93&\textbf{89.21}&90.86&84.66&87.61&\textbf{71.66}&68.23&\textbf{78.58}&\textbf{78.20}&65.55&79.15& 12.7    \\ \hline

    \end{tabular}%
    }
    \end{center}
    \label{tab:sota}
   \vspace{-3mm}
\end{table*}

\begin{table*}[t!]
  \caption{Comparisons of state-of-the-art methods on HRSC2016. \#Anchor means the number of anchors at each location of the feature map. $^*$ indicates that the result is evaluated under PASCAL VOC2012 metrics.}
  \vspace{-7mm}
   \begin{center}
    \resizebox{\textwidth}{!}{%
    \begin{tabular}{c|cccccccc|c} \hline
        Methods  & RC2~\cite{liu2017rrpnship}  & R$^2$PN$^*$~\cite{zhang2018toward} & RRD~\cite{liao2018rotation}  & RoI Trans.~\cite{ding2018transformer} & Xu \etal~\cite{xu2019gliding} & R$^3$Det~\cite{yang2019r3det} & DRN~\cite{pan2020dynamic} & CenterMap-Net~\cite{wang2020centermap}  & S$^2$A-Net (Ours) \\ \hline
        \#Anchor & -                           & 24                                 & 13                           & 20                                    & 20                            & 21                               & -                         & 15                                  & \textbf{1} \\ \hline
        mAP      & 75.7                       & 79.6                                & 84.3                         & 86.2                                  & 88.2                          & 89.26                             & 92.7$^*$                  & 92.8$^*$                            & \textbf{90.17} / \textbf{95.01$^*$} \\ \hline
        \end{tabular}
    }    
\end{center}

    \label{tab:hrsc2016}
  \vspace{-3mm}
\end{table*}

\begin{figure*}[t!]
  \centering
  \includegraphics[width= .95\textwidth]{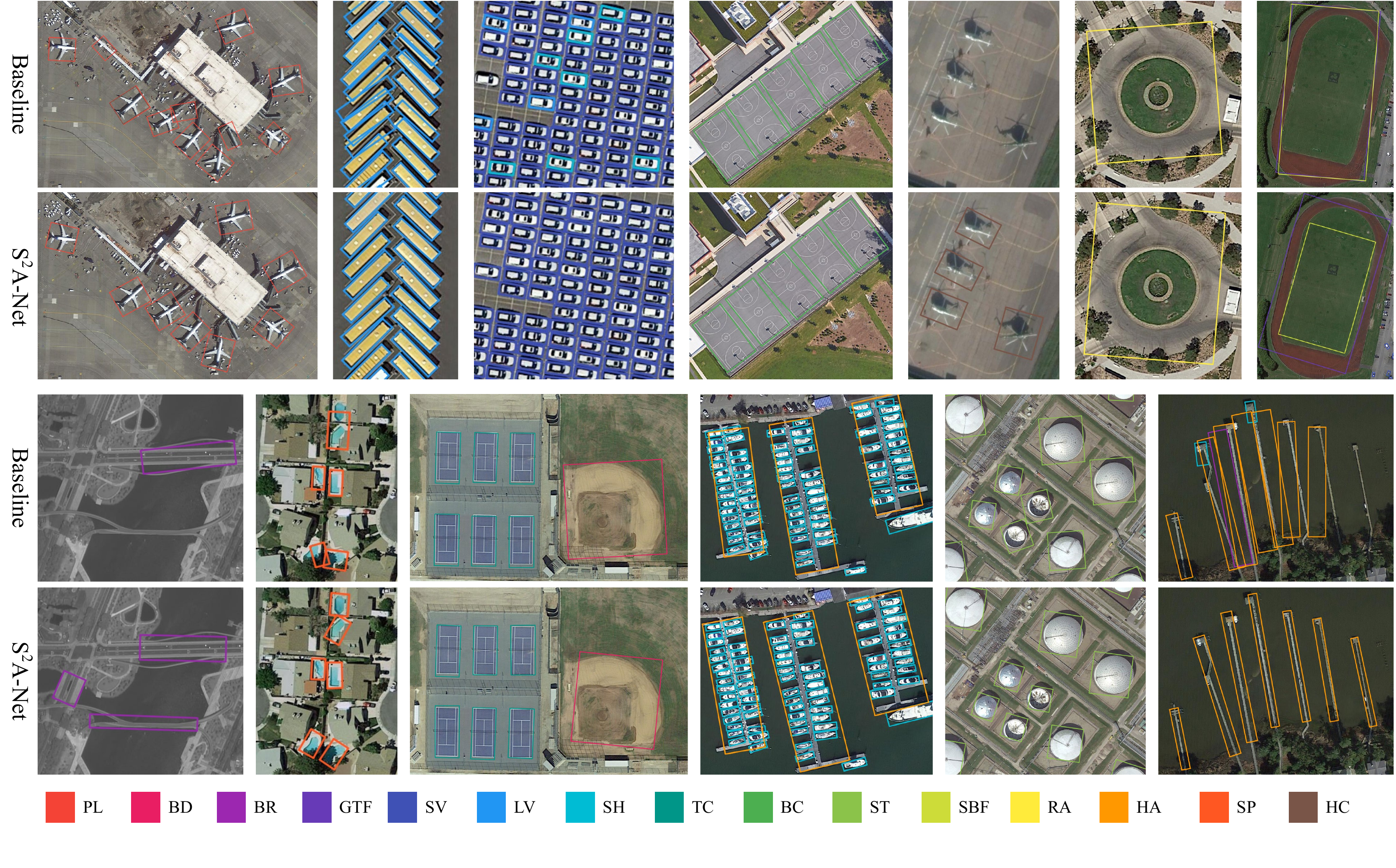}
  \vspace{-3mm}
  \caption{Some detection results on DOTA by different methods. For each image pair, the upper is the baseline method while the bottom is by S$^2$A-Net.}
  \label{fig:dota_results}
  \vspace{-3mm}
\end{figure*}

\begin{figure}[t!]
  \centering
  \includegraphics[width= .9\linewidth]{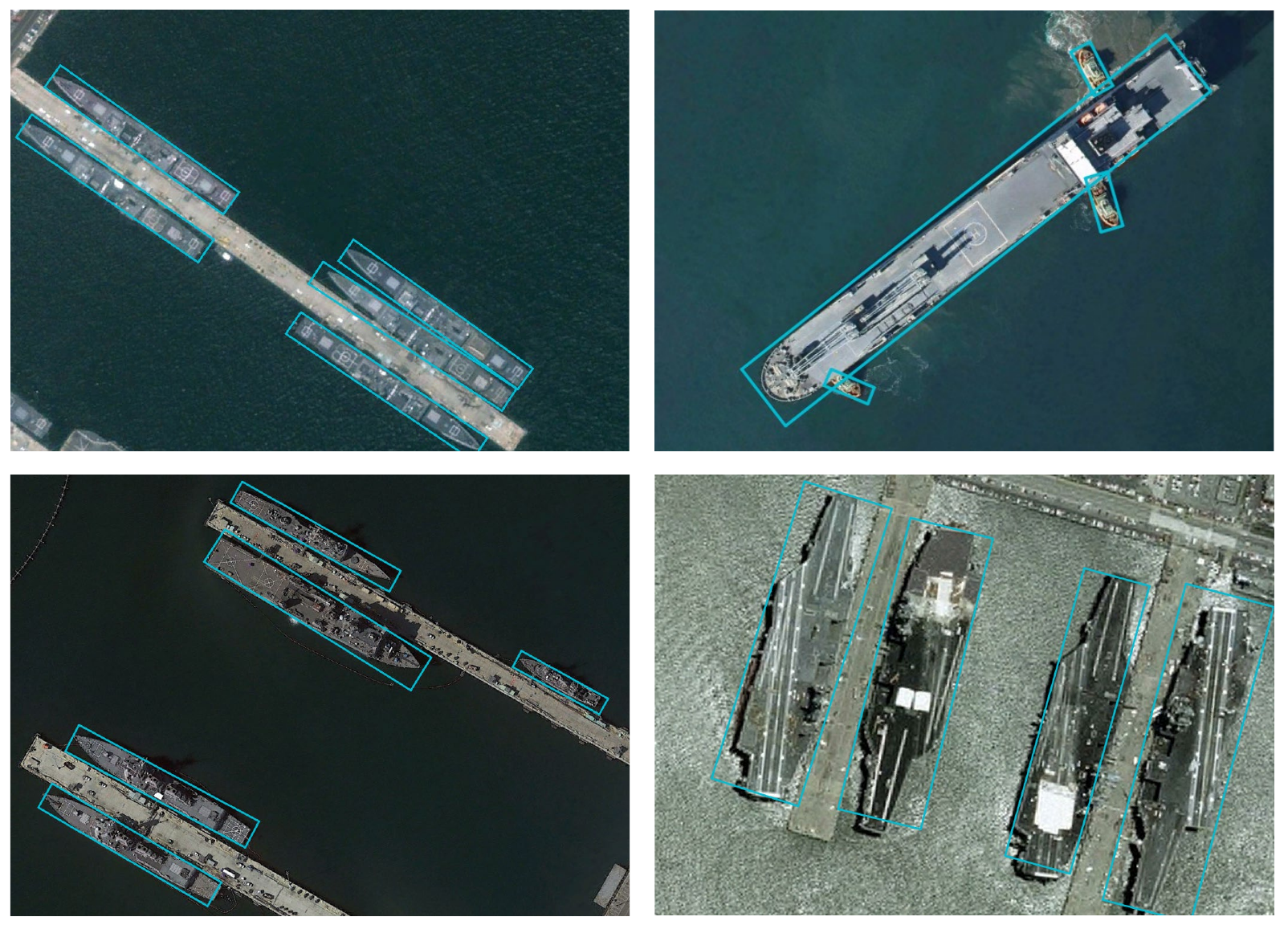}
  \vspace{-3mm}
  \caption{Some detection results on HRSC2016 with the proposed S$^2$A-Net.}
  \label{fig:hrsc2016_results}
  \vspace{-4mm}
\end{figure}

\subsection{Comparisons with the State-of-the-Art}
In this section, we compare our proposed S$^2$A-Net with other state-of-the-art methods on two aerial detection datasets, DOTA and HRSC2016. The settings have been introduced in Section~\ref{sec:dataset} and Section~\ref{sec:impl_detail}. 

{\bf Results on DOTA\footnote{The result is available at~\url{https://captain-whu.github.io/DOTA/results.html} with setting name \texttt{hanjiaming}. Note that, to concentrate on studying the algorithmic problem of ODAI, this setting is without using model fusions which can further improve the detection performance.}.} 
Note that RetinaNet is our re-implemented version referred in Sec~\ref{sec:retinanet}. As shown in Table~\ref{tab:sota}, we achieve 74.01\% mAP in 22.6 FPS with ResNet-50-FPN backbone and without any data augmentation (\eg, random rotation). Note that the FPS is an average FPS and we obtain it by calculating the overall inference time and the number of chip images (\ie, 10833). Besides, we achieve state-of-the-art 76.11\% mAP with a ResNet101 FPN backbone, outperforming all two-stage and one-stage methods. In multi-scale experiments, our S$^2$A-Net achieves 79.42\% and 79.15\% mAP with a ResNet-50-FPN and ResNet-101-FPN backbone, respectively. And we achieve best results in 10/15 categories, especially for some hard categories (\eg, bridge, soccer-ball field, swimming pool, helicopter). Qualitative detection results of the baseline method (\ie, RetinaNet) and our S$^2$A-Net are visualized in Fig~\ref{fig:dota_results}. Compared with RetinaNet, our S$^2$A-Net produces less false predictions when detecting on the object with dense distribution and large scale variations.

{\bf Results on HRSC2016.} Note that DRN~\cite{pan2020dynamic} and CenterMap-Net~\cite{wang2020centermap} are evaluated under PASCAL VOC2012 metrics while other methods are evaluated under PASCAL VOC2007 metrics, and the performance under VOC2012 metrics is better than that under VOC2007 metrics. As shown in Table~\ref{tab:hrsc2016}, our proposed S$^2$A-Net achieves 90.17\% and 95.01\% mAP under VOC2007 and VOC2012 metrics respectively, outperforming all other methods. The objects in HRSC2016 have large aspect ratios and arbitrary orientations, and previous methods often set more anchors for better performance, \eg, 20 in RoI Trans. and 21 in R$^3$Det. Compared with the previous best result 89.26\% (VOC2007) by R$^3$Det and 92.8\% (VOC2012) by CenterMap-Net, we improve 0.91\% and 2.21\% mAP respectively with only one anchor, which effectively get ride of heuristically defined anchors. Some qualitative results are shown in Fig.~\ref{fig:hrsc2016_results}.

\section{Conclusion}
\label{sec:conclusion}
In this paper, we propose a simple and effective Single-Shot Alignment Network (S$^2$A-Net) for oriented object detection in aerial images. With the proposed Feature Alignment Module and Oriented Detection Module, our S$^2$A-Net realizes full feature alignment and alleviates the inconsistency between regression ans classification. Besides, we explore the approach to detect on large-size images for better speed-accuracy trade-off. Extensive experiments demonstrate that our S$^2$A-Net can achieve state-of-the-art performance on both DOTA and HRSC2016.

\nocite{*}
{
\bibliographystyle{IEEEtran}
\small
\bibliography{main}
}

\end{document}